\documentclass{bmvc2k}

\usepackage{comment}

\usepackage{bm}
\usepackage{amssymb}
\usepackage{multirow}
\usepackage[capitalize]{cleveref}
\crefname{section}{Sec.}{Sec.}
\usepackage{subcaption}
\usepackage{adjustbox}
\usepackage{booktabs}
\usepackage[svgnames,table]{xcolor}

\newcommand{\punctspace}{\ }

\newcommand{\func}[2]{#1\mathopen{}\left(#2\mathclose{}\right)}

\newcommand{\pmf}[2][]{\func{P_{#1}}{#2}}

\newcommand{\upmf}[2][]{\func{\tilde{P}_{#1}}{#2}}
\newcommand{\given}{\mid}

\newcommand{\energyf}[2][]{\func{E_{#1}}{#2}}

\makeatletter

\newcommand{\minimize}{\@ifstar{\@minimizes}{\@minimize}}
\newcommand{\@minimizes}[1]{\ensuremath{ \operatorname{minimize } } }
\newcommand{\@minimize }[1]{\ensuremath{&\operatorname{minimize} &&}}
\newcommand{\maximize}{\@ifstar{\@maximizes}{\@maximize}}
\newcommand{\@maximizes}[1]{\ensuremath{ \operatorname{maximize } } }
\newcommand{\@maximize }[1]{\ensuremath{&\operatorname{maximize} &&}}
\newcommand{\minimizex}{\@ifstar{\@minimizexs}{\@minimizex}}
\newcommand{\@minimizexs}[1]{\ensuremath{ \underset{#1}{\operatorname{minimize}}\ }}
\newcommand{\@minimizex }[1]{\ensuremath{&\underset{#1}{\operatorname{minimize}}&&}}
\newcommand{\maximizex}{\@ifstar{\@maximizexs}{\@maximizex}}
\newcommand{\@maximizexs}[1]{\ensuremath{ \underset{#1}{\operatorname{maximize}}\ }}
\newcommand{\@maximizex }[1]{\ensuremath{&\underset{#1}{\operatorname{maximize}}&&}}
\newcommand{\argmin}{\@ifstar{\@argmins}{\@argmin}}
\newcommand{\@argmins}[1]{\ensuremath{ \operatorname{argmin } } }
\newcommand{\@argmin }[1]{\ensuremath{&\operatorname{argmin} &&}}
\newcommand{\argmax}{\@ifstar{\@argmaxs}{\@argmax}}
\newcommand{\@argmaxs}[1]{\ensuremath{ \operatorname{argmax } } }
\newcommand{\@argmax }[1]{\ensuremath{&\operatorname{argmax} &&}}
\newcommand{\argminx}{\@ifstar{\@argminxs}{\@argminx}}
\newcommand{\@argminxs}[1]{\ensuremath{ \underset{#1}{\operatorname{argmin}}\ }}
\newcommand{\@argminx }[1]{\ensuremath{&\underset{#1}{\operatorname{argmin}}&&}}
\newcommand{\argmaxx}{\@ifstar{\@argmaxxs}{\@argmaxx}}
\newcommand{\@argmaxxs}[1]{\ensuremath{ \underset{#1}{\operatorname{argmax}}\ }}
\newcommand{\@argmaxx }[1]{\ensuremath{&\underset{#1}{\operatorname{argmax}}&&}}

\makeatother

\renewcommand{\vec}[1]{\ensuremath{\bm{\MakeLowercase{#1}}}}

\newcommand{\imgsymb}{I}

\newcommand{\col}{u}
\newcommand{\row}{v}

\makeatletter

\newcommand{\stixelsymb}{s}
\newcommand{\stixelidx}{i}

\newcommand{\stixel}{\@ifstar{\@stixels}{\@stixel}}
\newcommand{\@stixels}[1]{ \bm{\MakeLowercase{\stixelsymb}}_{#1}  }
\newcommand{\@stixel }[1]{ \bm{\MakeUppercase{\stixelsymb}}_{#1}  }

\newcommand{\stixeld}{\@ifstar{\@stixelds}{\@stixeld}}
\newcommand{\@stixelds}{ \stixel*{\stixelidx}  }
\newcommand{\@stixeld }{ \stixel {\stixelidx}  }

\newcommand{\stixelu}{\@ifstar{\@stixelus}{\@stixelu}}
\newcommand{\@stixelus}[1]{ \stixel*{\col #1}   }
\newcommand{\@stixelu }[1]{ \stixel {\col #1}   }

\newcommand{\stixelud}{\@ifstar{\@stixeluds}{\@stixelud}}
\newcommand{\@stixeluds}{ \stixelu*{\stixelidx}  }
\newcommand{\@stixelud }{ \stixelu {\stixelidx}  }

\newcommand{\stixelcol}{\@ifstar{\@stixelcols}{\@stixelcol}}
\newcommand{\@stixelcols}{ \bm{\MakeLowercase{\stixelsymb}}_{:}  }
\newcommand{\@stixelcol }{ \bm{\MakeUppercase{\stixelsymb}}_{:}  }

\newcommand{\stixelcolu}{\@ifstar{\@stixelcolus}{\@stixelcolu}}
\newcommand{\@stixelcolus}{ \bm{\MakeLowercase{\stixelsymb}}_{\col :}  }
\newcommand{\@stixelcolu }{ \bm{\MakeUppercase{\stixelsymb}}_{\col :}  }

\newcommand{\numstixelssymb}{n}

\newcommand{\numstixels}{\@ifstar{\@numstixelss}{\@numstixels}}
\newcommand{\@numstixelss}{ \MakeLowercase{\numstixelssymb}  }
\newcommand{\@numstixels }{ \MakeUppercase{\numstixelssymb}  }

\newcommand{\numstixelsu}{\@ifstar{\@numstixelsus}{\@numstixelsu}}
\newcommand{\@numstixelsus}{ {\numstixels*}_{\col}  }
\newcommand{\@numstixelsu }{ {\numstixels }_{\col}  }

\newcommand{\rowsymb}{v}
\newcommand{\rowtopsymb}{t}
\newcommand{\rowbotsymb}{b}

\newcommand{\rowtop}{\@ifstar{\@rowtops}{\@rowtop}}
\newcommand{\@rowtops}[1]{ \MakeLowercase{\rowsymb}_{#1}^{\text{\rowtopsymb}}  }
\newcommand{\@rowtop }[1]{ \MakeUppercase{\rowsymb}_{#1}^{\text{\rowtopsymb}}  }

\newcommand{\rowtopd}{\@ifstar{\@rowtopds}{\@rowtopd}}
\newcommand{\@rowtopds}{ \rowtop*{\stixelidx}  }
\newcommand{\@rowtopd }{ \rowtop {\stixelidx}  }

\newcommand{\rowtopu}{\@ifstar{\@rowtopus}{\@rowtopu}}
\newcommand{\@rowtopus}[1]{ \rowtop*{\col #1}  }
\newcommand{\@rowtopu }[1]{ \rowtop {\col #1}  }

\newcommand{\rowtopud}{\@ifstar{\@rowtopuds}{\@rowtopud}}
\newcommand{\@rowtopuds}{ \rowtopu*{\stixelidx}  }
\newcommand{\@rowtopud }{ \rowtopu {\stixelidx}  }

\newcommand{\rowbot}{\@ifstar{\@rowbots}{\@rowbot}}
\newcommand{\@rowbots}[1]{ \MakeLowercase{\rowsymb}_{#1}^{\text{\rowbotsymb}}  }
\newcommand{\@rowbot }[1]{ \MakeUppercase{\rowsymb}_{#1}^{\text{\rowbotsymb}}  }

\newcommand{\rowbotd}{\@ifstar{\@rowbotds}{\@rowbotd}}
\newcommand{\@rowbotds}{ \rowbot*{\stixelidx}  }
\newcommand{\@rowbotd }{ \rowbot {\stixelidx}  }

\newcommand{\rowbotu}{\@ifstar{\@rowbotus}{\@rowbotu}}
\newcommand{\@rowbotus}[1]{ \rowbot*{\col #1}  }
\newcommand{\@rowbotu }[1]{ \rowbot {\col #1}  }

\newcommand{\rowbotud}{\@ifstar{\@rowbotuds}{\@rowbotud}}
\newcommand{\@rowbotuds}{ \rowbotu*{\stixelidx}  }
\newcommand{\@rowbotud }{ \rowbotu {\stixelidx}  }

\newcommand{\classsymb}{c}

\newcommand{\groundsymb}{g}
\newcommand{\ground}{\mathcal{\MakeUppercase{\groundsymb}}}
\newcommand{\objectsymb}{o}
\newcommand{\object}{\mathcal{\MakeUppercase{\objectsymb}}}
\newcommand{\skysymb}{s}
\newcommand{\sky}{\mathcal{\MakeUppercase{\skysymb}}}

\newcommand{\class}{\@ifstar{\@classs}{\@class}}
\newcommand{\@classs}[1]{ \MakeLowercase{\classsymb}_{#1}  }
\newcommand{\@class }[1]{ \MakeUppercase{\classsymb}_{#1}  }

\newcommand{\classd}{\@ifstar{\@classds}{\@classd}}
\newcommand{\@classds}{ \class*{\stixelidx}  }
\newcommand{\@classd }{ \class {\stixelidx}  }

\newcommand{\classu}{\@ifstar{\@classus}{\@classu}}
\newcommand{\@classus}[1]{ \class*{\col #1}  }
\newcommand{\@classu }[1]{ \class {\col #1}  }

\newcommand{\classud}{\@ifstar{\@classuds}{\@classud}}
\newcommand{\@classuds}{ \classu*{\stixelidx}  }
\newcommand{\@classud }{ \classu {\stixelidx}  }

\newcommand{\dispattrsymb}{d}

\newcommand{\dispattr}{\@ifstar{\@dispattrs}{\@dispattr}}
\newcommand{\@dispattrs}[1]{ \MakeLowercase{\dispattrsymb}_{#1}  }
\newcommand{\@dispattr }[1]{ \MakeUppercase{\dispattrsymb}_{#1}  }

\newcommand{\dispattrd}{\@ifstar{\@dispattrds}{\@dispattrd}}
\newcommand{\@dispattrds}{ \dispattr*{\stixelidx}  }
\newcommand{\@dispattrd }{ \dispattr {\stixelidx}  }

\newcommand{\dispattru}{\@ifstar{\@dispattrus}{\@dispattru}}
\newcommand{\@dispattrus}[1]{ \dispattr*{\col #1}  }
\newcommand{\@dispattru }[1]{ \dispattr {\col #1}  }

\newcommand{\dispattrud}{\@ifstar{\@dispattruds}{\@dispattrud}}
\newcommand{\@dispattruds}{ \dispattru*{\stixelidx}  }
\newcommand{\@dispattrud }{ \dispattru {\stixelidx}  }

\newcommand{\imgattrsymb}{o}

\newcommand{\imgattr}{\@ifstar{\@imgattrs}{\@imgattr}}
\newcommand{\@imgattrs}[1]{ \MakeLowercase{\imgattrsymb}_{#1}  }
\newcommand{\@imgattr }[1]{ \MakeUppercase{\imgattrsymb}_{#1}  }

\newcommand{\imgattrd}{\@ifstar{\@imgattrds}{\@imgattrd}}
\newcommand{\@imgattrds}{ \imgattr*{\stixelidx}  }
\newcommand{\@imgattrd }{ \imgattr {\stixelidx}  }

\newcommand{\imgattru}{\@ifstar{\@imgattrus}{\@imgattru}}
\newcommand{\@imgattrus}[1]{ \imgattr*{\col #1}  }
\newcommand{\@imgattru }[1]{ \imgattr {\col #1}  }

\newcommand{\imgattrud}{\@ifstar{\@imgattruds}{\@imgattrud}}
\newcommand{\@imgattruds}{ \imgattru*{\stixelidx}  }
\newcommand{\@imgattrud }{ \imgattru {\stixelidx}  }

\newcommand{\dispmeassymb}{D}

\newcommand{\disp}{\@ifstar{\@disps}{\@disp}}
\newcommand{\@disps}[1]{ \MakeLowercase{\dispmeassymb}_{#1}  }
\newcommand{\@disp }[1]{ \MakeUppercase{\dispmeassymb}_{#1}  }

\newcommand{\dispd}{\@ifstar{\@dispds}{\@dispd}}
\newcommand{\@dispds}{ \disp*{\row}  }
\newcommand{\@dispd }{ \disp {\row}  }

\newcommand{\dispu}{\@ifstar{\@dispus}{\@dispu}}
\newcommand{\@dispus}[1]{ \disp*{\col #1}  }
\newcommand{\@dispu }[1]{ \disp {\col #1}  }

\newcommand{\dispud}{\@ifstar{\@dispuds}{\@dispud}}
\newcommand{\@dispuds}{ \dispu*{\row}  }
\newcommand{\@dispud }{ \dispu {\row}  }

\newcommand{\dispcol}{\@ifstar{\@dispcols}{\@dispcol}}
\newcommand{\@dispcols}{ \bm{\MakeLowercase{\dispmeassymb}}_{:}  }
\newcommand{\@dispcol }{ \bm{\MakeUppercase{\dispmeassymb}}_{:}  }

\newcommand{\dispcolu}{\@ifstar{\@dispcolus}{\@dispcolu}}
\newcommand{\@dispcolus}{ \bm{\MakeLowercase{\dispmeassymb}}_{\col :}  }
\newcommand{\@dispcolu }{ \bm{\MakeUppercase{\dispmeassymb}}_{\col :}  }

\newcommand{\confmeassymb}{C}
\newcommand{\conf}{\@ifstar{\@confs}{\@conf}}
\newcommand{\@confs}[1]{ \MakeLowercase{\confmeassymb}_{#1}  }
\newcommand{\@conf }[1]{ \MakeUppercase{\confmeassymb}_{#1}  }

\newcommand{\confd}{\@ifstar{\@confds}{\@confd}}
\newcommand{\@confds}{ \conf*{\row}  }
\newcommand{\@confd }{ \conf {\row}  }

\newcommand{\imgval}{\@ifstar{\@imgvals}{\@imgval}}
\newcommand{\@imgvals}[1]{ \MakeLowercase{\imgsymb}_{#1}  }
\newcommand{\@imgval }[1]{ \MakeUppercase{\imgsymb}_{#1}  }

\newcommand{\imgvald}{\@ifstar{\@imgvalds}{\@imgvald}}
\newcommand{\@imgvalds}{ \imgval*{\row}  }
\newcommand{\@imgvald }{ \imgval {\row}  }

\newcommand{\imgvalu}{\@ifstar{\@imgvalus}{\@imgvalu}}
\newcommand{\@imgvalus}[1]{ \imgval*{\col #1}  }
\newcommand{\@imgvalu }[1]{ \imgval {\col #1}  }

\newcommand{\imgvalud}{\@ifstar{\@imgvaluds}{\@imgvalud}}
\newcommand{\@imgvaluds}{ \imgvalu*{\row}  }
\newcommand{\@imgvalud }{ \imgvalu {\row}  }

\newcommand{\imgcol}{\@ifstar{\@imgcols}{\@imgcol}}
\newcommand{\@imgcols}{ \bm{\MakeLowercase{\imgsymb}}_{:}  }
\newcommand{\@imgcol }{ \bm{\MakeUppercase{\imgsymb}}_{:}  }

\newcommand{\imgcolu}{\@ifstar{\@imgcolus}{\@imgcolu}}
\newcommand{\@imgcolus}{ \bm{\MakeLowercase{\imgsymb}}_{\col :}  }
\newcommand{\@imgcolu }{ \bm{\MakeUppercase{\imgsymb}}_{\col :}  }

\newcommand{\pixclasssymb}{L}

\newcommand{\pixclassval}{\@ifstar{\@pixclassvals}{\@pixclassval}}
\newcommand{\@pixclassvals}[1]{ \MakeLowercase{\pixclasssymb}_{#1}  }
\newcommand{\@pixclassval }[1]{ \MakeUppercase{\pixclasssymb}_{#1}  }

\newcommand{\pixclassvalvec}{\@ifstar{\@pixclassvalvecs}{\@pixclassvalvec}}
\newcommand{\@pixclassvalvecs}[1]{ \vec{\MakeLowercase{\pixclasssymb}}_{#1}  }
\newcommand{\@pixclassvalvec }[1]{ \vec{\MakeUppercase{\pixclasssymb}}_{#1}  }

\newcommand{\pixclassvald}{\@ifstar{\@pixclassvalds}{\@pixclassvald}}
\newcommand{\@pixclassvalds}{ \pixclassval*{\row}  }
\newcommand{\@pixclassvald }{ \pixclassval {\row}  }

\newcommand{\pixclassvaldvec}{\@ifstar{\@pixclassvaldvecs}{\@pixclassvaldvec}}
\newcommand{\@pixclassvaldvecs}{ \pixclassvalvec*{\row}  }
\newcommand{\@pixclassvaldvec }{ \pixclassvalvec {\row}  }

\newcommand{\pixclasscol}{\@ifstar{\@pixclasscols}{\@pixclasscol}}
\newcommand{\@pixclasscols}{ \bm{\MakeLowercase{\pixclasssymb}}_{:}  }
\newcommand{\@pixclasscol }{ \bm{\MakeUppercase{\pixclasssymb}}_{:}  }

\newcommand{\pixclasscolu}{\@ifstar{\@pixclasscolus}{\@pixclasscolu}}
\newcommand{\@pixclasscolus}{ \bm{\MakeLowercase{\pixclasssymb}}_{\col :}  }
\newcommand{\@pixclasscolu }{ \bm{\MakeUppercase{\pixclasssymb}}_{\col :}  }

\newcommand{\cutpriorsymb}{C}

\newcommand{\cutpriorval}{\@ifstar{\@cutpriorvals}{\@cutpriorval}}
\newcommand{\@cutpriorvals}[1]{ \MakeLowercase{\cutpriorsymb}_{#1}  }
\newcommand{\@cutpriorval }[1]{ \MakeUppercase{\cutpriorsymb}_{#1}  }

\newcommand{\cutpriorvalvec}{\@ifstar{\@cutpriorvalvecs}{\@cutpriorvalvec}}
\newcommand{\@cutpriorvalvecs}[1]{ \vec{\MakeLowercase{\cutpriorsymb}}_{#1}  }
\newcommand{\@cutpriorvalvec }[1]{ \vec{\MakeUppercase{\cutpriorsymb}}_{#1}  }

\newcommand{\cutpriorvald}{\@ifstar{\@cutpriorvalds}{\@cutpriorvald}}
\newcommand{\@cutpriorvalds}{ \cutpriorval*{\row}  }
\newcommand{\@cutpriorvald }{ \cutpriorval {\row}  }

\newcommand{\cutpriorvaldvec}{\@ifstar{\@cutpriorvaldvecs}{\@cutpriorvaldvec}}
\newcommand{\@cutpriorvaldvecs}{ \cutpriorvalvec*{\row}  }
\newcommand{\@cutpriorvaldvec }{ \cutpriorvalvec {\row}  }

\newcommand{\cutpriorcol}{\@ifstar{\@cutpriorcols}{\@cutpriorcol}}
\newcommand{\@cutpriorcols}{ \bm{\MakeLowercase{\cutpriorsymb}}_{:}  }
\newcommand{\@cutpriorcol }{ \bm{\MakeUppercase{\cutpriorsymb}}_{:}  }

\newcommand{\cutpriorcolu}{\@ifstar{\@cutpriorcolus}{\@cutpriorcolu}}
\newcommand{\@cutpriorcolus}{ \bm{\MakeLowercase{\cutpriorsymb}}_{\col :}  }
\newcommand{\@cutpriorcolu }{ \bm{\MakeUppercase{\cutpriorsymb}}_{\col :}  }

\newcommand{\meassymb}{M}

\newcommand{\meas}{\@ifstar{\@meass}{\@meas}}
\newcommand{\@meass}[1]{ \bm{\MakeLowercase{\meassymb}}_{#1}  }
\newcommand{\@meas }[1]{ \bm{\MakeUppercase{\meassymb}}_{#1}  }

\newcommand{\meascol}{\@ifstar{\@meascols}{\@meascol}}
\newcommand{\@meascols}{ \bm{\MakeLowercase{\meassymb}}_{:}  }
\newcommand{\@meascol }{ \bm{\MakeUppercase{\meassymb}}_{:}  }

\newcommand{\meascolu}{\@ifstar{\@meascolus}{\@meascolu}}
\newcommand{\@meascolus}{ \bm{\MakeLowercase{\meassymb}}_{\col :}  }
\newcommand{\@meascolu }{ \bm{\MakeUppercase{\meassymb}}_{\col :}  }

\newcommand{\pixmeasd}{\@ifstar{\@pixmeasds}{\@pixmeasd}}
\newcommand{\@pixmeasds}{ \meas*{\row}  }
\newcommand{\@pixmeasd }{ \meas {\row}  }

\newcommand{\priorcoltop}{\@ifstar{\@priorcoltops}{\@priorcoltop}}
\newcommand{\@priorcoltops}{ \bm{\MakeLowercase{\rowtopsymb}}_{:}  }
\newcommand{\@priorcoltop }{ \bm{\MakeUppercase{\rowtopsymb}}_{:}  }

\newcommand{\priorcoltopu}{\@ifstar{\@priorcoltopus}{\@priorcoltopu}}
\newcommand{\@priorcoltopus}{ \bm{\MakeLowercase{\rowtopsymb}}_{\col :}  }
\newcommand{\@priorcoltopu }{ \bm{\MakeUppercase{\rowtopsymb}}_{\col :}  }

\newcommand{\priorcolbot}{\@ifstar{\@priorcolbots}{\@priorcolbot}}
\newcommand{\@priorcolbots}{ \bm{\MakeLowercase{\rowbotsymb}}_{:}  }
\newcommand{\@priorcolbot }{ \bm{\MakeUppercase{\rowbotsymb}}_{:}  }

\newcommand{\priorcolbotu}{\@ifstar{\@priorcolbotus}{\@priorcolbotu}}
\newcommand{\@priorcolbotus}{ \bm{\MakeLowercase{\rowbotsymb}}_{\col :}  }
\newcommand{\@priorcolbotu }{ \bm{\MakeUppercase{\rowbotsymb}}_{\col :}  }

\newcommand{\partitionfunction}{Z}

\newcommand{\proboutlierdisp}{p_{\text{out}}}

\newcommand{\eprior}[1]{\energyf[prior]{#1}}
\newcommand{\edata}[1]{\energyf[data]{#1}}
\newcommand{\edepthmodel}[1]{\energyf[plane]{#1}}

\newcommand{\depthmodelsymb}{\mu}
\newcommand{\depthmodel}{\func{\depthmodelsymb{}}{\stixeld*,\row}}

\newcommand{\planeasymb}{a}
\newcommand{\planebsymb}{b}

\newcommand{\planea}{\@ifstar{\@planeas}{\@planea}}
\newcommand{\@planeas}{ \MakeLowercase{\planeasymb} }
\newcommand{\@planea }{ \MakeUppercase{\planeasymb} }

\newcommand{\planeb}{\@ifstar{\@planebs}{\@planeb}}
\newcommand{\@planebs}{ \MakeLowercase{\planebsymb} }
\newcommand{\@planeb }{ \MakeUppercase{\planebsymb} }

\makeatother

\usepackage{scalerel}

\usepackage{tikz}
\usetikzlibrary{calc}
\usetikzlibrary{shapes}
\usetikzlibrary{fit}
\usetikzlibrary{chains}
\usetikzlibrary{arrows}
\usetikzlibrary{decorations.pathreplacing}
\usetikzlibrary{patterns}
\usetikzlibrary{intersections}

\usepackage{pgfplots}
\pgfplotsset{compat=newest}
\pgfplotsset{plot coordinates/math parser=false}
\usetikzlibrary{pgfplots.groupplots}

\usepackage{tabularx,ragged2e}

\newcolumntype{L}{>{\raggedright\arraybackslash}X}

\definecolor{sidewalk}{rgb}{0.953125,0.13671875,0.90625}
\definecolor{road}{rgb}{0.5,0.25,0.5}
\definecolor{traffic light}{rgb}{0.9765625,0.6640625,0.1171875}
\definecolor{traffic sign}{rgb}{0.859375,0.859375,0.}
\definecolor{vegetation}{rgb}{0.41796875,0.5546875,0.13671875}
\definecolor{person}{rgb}{0.859375,0.078125,0.234375}
\definecolor{car}{rgb}{0.,0.,0.5546875}
\definecolor{fence}{rgb}{0.7421875,0.59765625,0.59765625}

\definecolor{terrain}{rgb}{0.59375,0.98046875,0.59375}
\definecolor{building}{rgb}{0.2734375,0.2734375,0.2734375}
\definecolor{wall}{rgb}{0.3984375,0.3984375,0.609375}
\definecolor{rider}{rgb}{0.99609375,0.,0.}
\definecolor{truck}{rgb}{0.,0.,0.2734375}
\definecolor{bus}{rgb}{0.,0.234375,0.390625}
\definecolor{train}{rgb}{0.,0.3125,0.390625}
\definecolor{motorcycle}{rgb}{0.,0.,0.8984375}
\definecolor{bicycle}{rgb}{0.46484375,0.04296875,0.125}
\definecolor{sky}{rgb}{0.2734375,0.5078125,0.703125}
\definecolor{pole}{rgb}{0.5078125,0.5078125,0.5078125}
\definecolor{road lines}{rgb}{0.615686274509804,0.9176470588235294,0.19607843137254902}

\newcommand\labelcolor[1]{\textcolor{white}{\cellcolor{#1}{\scriptsize #1}}}
\newcommand\labelcolorb[1]{\cellcolor{#1}{\scriptsize #1}}

\title{Slanted Stixels: Representing \\San Francisco's Steepest Streets}

\addauthor{Daniel Hernandez-Juarez*$\dagger$}{http://www.cvc.uab.es/people/dhernandez/}{1}
\addauthor{Lukas Schneider*}{lukas.schneider@daimler.com}{3}
\addauthor{Antonio Espinosa}{antoniomiguel.espinosa@uab.es}{1}
\addauthor{David V{\'a}zquez\textsuperscript{1,}}{dvazquez@cvc.uab.es}{2}
\addauthor{Antonio M. L{\'o}pez\textsuperscript{1,}}{antonio@cvc.uab.es}{2}
\addauthor{Uwe Franke}{uwe.franke@daimler.com}{3}
\addauthor{Marc Pollefeys}{marc.pollefeys@inf.ethz.ch}{4}
\addauthor{Juan C. Moure}{juancarlos.moure@uab.es}{1}

\addinstitution{
 Universitat Aut{\`o}noma de Barcelona\\
 Barcelona, Spain
}
\addinstitution{
 Computer Vision Center\\
 Barcelona, Spain
}
\addinstitution{
 Daimler AG, R\&D\\
 B{\"o}blingen, Germany
}
\addinstitution{
 ETH Z{\"u}rich\\
 Z{\"u}rich, Switzerland
}

\runninghead{Daniel Hernandez-Juarez, Lukas Schneider}{Slanted Stixels}

\def\eg{\emph{e.g}\bmvaOneDot}
\def\ie{\emph{i.e}\bmvaOneDot}
\def\cf{\emph{c.f}\bmvaOneDot}
\def\Eg{\emph{E.g}\bmvaOneDot}

\def\datasetname{SYNTHIA-SF}
\def\datasetnamefull{SYNTHIA-San Francisco}

\begin{document}

\maketitle

\begin{abstract}
In this work we present a novel compact scene representation based on Stixels that infers geometric and semantic information. Our approach overcomes the previous rather restrictive geometric assumptions for Stixels by introducing a novel depth model to account for non-flat roads and slanted objects. Both semantic and depth cues are used jointly to infer the scene representation in a sound global energy minimization formulation. Furthermore, a novel approximation scheme is introduced that uses an extremely efficient over-segmentation. In doing so, the computational complexity of the Stixel inference algorithm is reduced significantly, achieving real-time computation capabilities with only a slight drop in accuracy. We evaluate the proposed approach in terms of semantic and geometric accuracy as well as run-time on four publicly available benchmark datasets. Our approach maintains accuracy on flat road scene datasets while improving substantially on a novel non-flat road dataset.
\end{abstract}

\section{Introduction}
\label{sec:intro}
Autonomous vehicles, advanced driver assistance systems, robots and other intelligent devices need to understand their environment; this requires both geometric (distance) and semantic (classification) information. This data must be represented in a very compact model and must be computed in real-time that can serve as building block of higher-level modules, such as localization and planning.

\begin{figure}[h]
\centering
\includegraphics[width=0.9\linewidth]{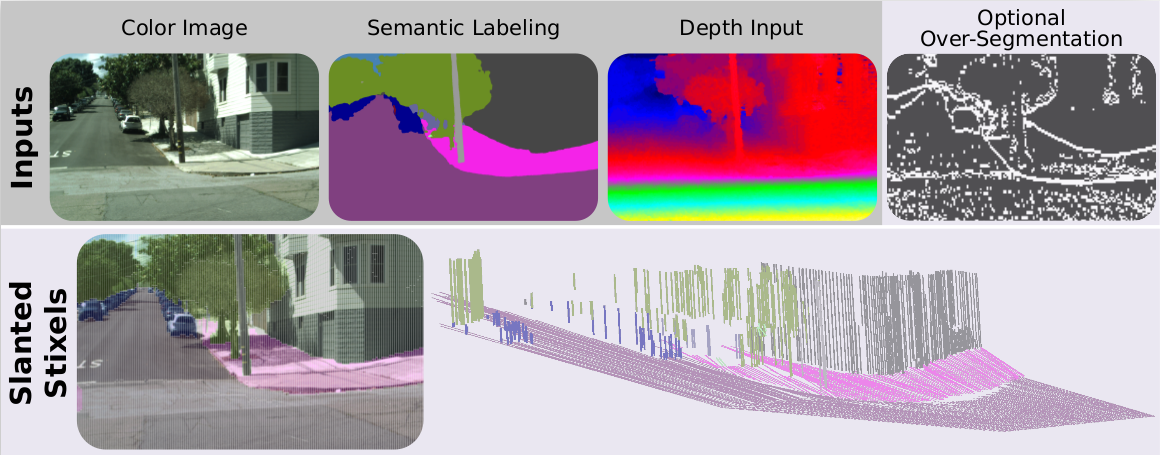}
\captionsetup{skip=\dimexpr\abovecaptionskip+10pt}
\caption{The proposed approach: pixel-wise semantic and depth information serve as input to our Slanted Stixels, a compact semantic 3D scene representation that accurately handles arbitrary scenarios as \eg San Francisco. The optional over-segmentation in the top-right yields significant speed gains nearly retaining the depth and semantic accuracy.
}
\label{fig:overview}
\end{figure}

The Stixel World has been successfully used for representing traffic scenes, as introduced in \cite{Pfeiffer2011}. The intelligent vehicles community has shown an increasing interest in this model over the last years \cite{Schneider2016, Hernandez2017, Benenson2011, Cordts2014, Cordts2017, Oana2016, Levi2015, carrillo2016}. It defines a compact medium-level representation of dense 3D disparity data obtained from stereo vision using rectangles (Stixels) as elements. These are classified either as \textit{ground}-like planes, upright \textit{objects} or \textit{sky}, which are the geometric primitives found in man-made environments. This converts millions of disparity pixels to hundreds or thousands of Stixels. At the same time, most task-relevant scene structures, such as free space and obstacles, are adequately represented.

A recent work \cite{Schneider2016} fuses geometric and semantic information in an extended Stixel model, which is able to provide a richer yet compact representation of the traffic scene. Our work extends \cite{Schneider2016} by incorporating a new depth model that takes arbitrary kinds of slanted objects and non-flat roads into account. The induced extra computational complexity is reduced in this paper by incorporating an over-segmentation strategy
that nearly retains the accuracy and can be applied to any Stixel model proposed so far.

Our work yields an improved Stixel representation that accounts for non-flat roads, outperforming the original Stixel model in this context while keeping the same accuracy on flat road scenes. An overview of our approach is shown in \cref{fig:overview}.


\section{Related work}
\label{sec:related_work}

Our proposed approach introduces a novel Stixel-based scene representation that is able to account for non-flat roads, \cf \cref{fig:overview}. We also devise an approximation to compute Stixels faster. Therefore, we see three categories of related publications.

The first group is comprised by road scene models. In most cases, occupancy grid maps are used to represent the surrounding of the vehicle~\cite{Dhiman2014,Muffert2014,Nuss2015,Thrun2002}. Typically a grid in bird's eye perspective is defined and used to detect occupied grid cells. These grids and the Stixel World both represent the 2D image in terms of column-wise stripes allowing to capture the camera data in a polar fashion. However, the Stixel inference in the image domain differs significantly from classical grid-based approaches.

The second category includes different Stixel-based methods. Stixels were originally devised to represent the 3D scene as observed by stereoscopic \cite{Pfeiffer2011, Benenson2011} or monocular imagery \cite{Levi2015}.
%
Our proposal is based on \cite{Schneider2016}: they use semantic cues in addition to depth to extract a Stixel representation.
However, they are limited to flat road scenarios due to constant height assumption.
In contrast, our proposal overcomes this drawback by incorporating a novel plane model together with effective priors on the plane parameters.

Finally, the third category consists of fast methods for Stixel computation. Some methods \cite{Badino2009, Benenson2011, Oana2016} model the scene with a single Stixel per column: they can be faster but provide an incomplete world model, \eg they cannot represent a pedestrian and a building in the same column. A recent work \cite{carrillo2016} uses edge-based disparity maps to compute Stixels: this method is also fast but gives inferior accuracy compared to the original Stixels \cite{pfeiffer2013exploiting}. The FPGA implementation from \cite{Muffert2014} runs at 25 Hz. Finally, \cite{Hernandez2017} present an embedded GPU implementation that runs at 26 Hz for Stixel widths of 5 px
computed using an SGM stereo algorithm also implemented on GPU \cite{Hernandez2016}. In contrast, we propose a novel algorithmic approximation that is hardware agnostic. Accordingly, it could also benefit of the aforementioned approaches.

Our main contributions are: (1) a real-time, compact and robust Stixel representation that incorporates a novel depth model to accurately represent arbitrary kinds of slanted surfaces in a sound probabilistic formulation; (2) a novel over-segmentation input to the main Stixel segmentation algorithm that significantly reduces the run-time of the method while nearly retaining its accuracy; (3) a new challenging synthetic dataset with non-flat roads that includes pixel-level semantic and depth ground-truth and allows to evaluate the accuracy of competing algorithms in such scenarios. This dataset will be made publicly available\footnotemark with this paper; (4) an in-depth evaluation in terms of run-time as well as semantic and depth accuracy carried out on this novel dataset and several real-world benchmarks. Compared to the existing state-of-the-art approach, the depth accuracy is substantially improved especially in non-flat road scenarios.
\footnotetext{http://synthia-dataset.net}

\section{Stixel Model}
\label{sec:method}



The Stixel world is a segmentation of image columns into stick-like super-pixels with class labels and a 3D planar depth model. This joint segmentation and labeling problem is carried out via optimization of the column-wise posterior distribution $\pmf{\stixel{:} \given \meas{:} }$ defined over a Stixel segmentation $\stixel{:} $ given all measurements $\meas{:} $ from that particular image column. In the following, we drop the column indexes for ease of notation. We obtain Stixel widths $>1$ as illustrated \eg in \cref{fig:overview} by down-sampling of the inputs.

A Stixel column segmentation consists of an arbitrary number $\numstixels$ of Stixels $\stixel{i}$, each representing four random variables: the Stixel extent via bottom $\rowbotd$ and top $\rowtopd$ row, as well as it's class $\classd$ and depth model $\dispattrd$. Thereby, the number of Stixels itself is a random variable that is optimized jointly during inference.
%
%
To this end, the posterior probability is defined by means of the unnormalized prior and likelihood distributions
${
  \pmf{\stixel{} \given \meas{}} =
        \frac{1}{\partitionfunction} \upmf{\meas{} \given \stixel{}} \upmf{\stixel{}} \punctspace
}$
transformed to log-likelihoods via $\pmf{\stixel{}=\stixel*{} \given \meas{}=\meas*{}} \allowbreak = -\log(e^{ - \energyf{\stixel*{},\meas*{}}})$.

The \textbf{likelihood} term $\edata{\cdot}$ thereby rates how well the measurements $\pixmeasd*$ at pixel $\row$ fit to the overlapping Stixel $\stixeld*$
\begin{equation}
\edata{\stixel*{},\meas*{}} =
\sum_{\stixelidx=1}^{\numstixels}
\energyf[stixel]{\stixeld*,\meas*{}}
=
\sum_{\stixelidx=1}^{\numstixels}
\sum_{\row = \rowbotd*}^{\rowtopd*}
\energyf[pixel]{\stixeld*,\pixmeasd*} \punctspace .
\label{eq:data}
\end{equation}
This pixel-wise energy is further split in a semantic and a depth term
\begin{equation}
\energyf[pixel]{\stixeld*,\pixmeasd*} =
\energyf[disp]{\stixeld*,\dispd*} + w_l \cdot \energyf[sem]{\stixeld*,\pixclassvald*} .
\end{equation}
The semantic energy favors semantic classes of the Stixel that fit to the observed pixel-level semantic input \cite{Schneider2016}. The parameter $w_l$ controls the influence of the semantic data term. The depth term is defined by means of a probabilistic and generative sensor model $\pmf[\row]{\cdot}$ that considers the accordance of the depth measurement $\dispd*$ at row $\row$ to the Stixel $\stixeld*$
\begin{equation}
\energyf[disp]{\stixeld*,\dispd*} =
- \func{\log}{\pmf[\row]{\dispd=\dispd* \given \stixeld=\stixeld*}} \punctspace .
\label{eq:data_depth}
\end{equation}
It is comprised of a constant outlier probability $\proboutlierdisp$ and a Gaussian
sensor noise model for valid measurements
with confidence $\confd*$
%
\begin{equation}
\pmf[\row]{\dispd \given \stixeld} = \frac{\proboutlierdisp}{\partitionfunction_{U}}
  + \frac{1-\proboutlierdisp}{\func{\partitionfunction_{G}}{\stixeld*}} e^{ -
    \left(
        \frac{ \confd*  \left(\dispd*-\depthmodel\right)}{\func{\sigma}{\stixeld*}}
    \right)^2 }
\label{eq:dispmeasmodel2}
\end{equation}
that is centered at the expected disparity $\depthmodel$ given the depth model of the Stixel.
$\partitionfunction_{U}$ and $\func{\partitionfunction_{G}}{\stixeld*}$ normalize the distributions.
%

This paper introduces a new plane depth model that overcomes the previous rather restrictive constant depth and constant height assumptions for \textit{object} respectively \textit{ground} Stixels.
To this end, we formulate the depth model $\depthmodel$ using two random variables defining a plane in the disparity space
that evaluates to the disparity in row $\row$ via
%
\begin{equation}
\depthmodel = \planeb*_\stixelidx \cdot \row + \planea*_\stixelidx \punctspace .
\end{equation}
Note that we assume narrow Stixels and thus can neglect one plane parameter, \ie the roll.

The \textbf{prior} captures knowledge about the segmentation: the Stixel segmentation has to be consistent, \ie each pixel is assigned to exactly one Stixel. A model complexity term favors solutions composed of fewer Stixels by invoking costs for each Stixel in the column segmentation $\stixel{}$. Furthermore, prior assumptions as "objects are likely to stand on the ground" and "sky is unlikely below the road surface" are taken into account. The interested reader is referred to \cite{Cordts2017} for more details. The Markov property is used so that the prior reduces to pair-wise relations between subsequent Stixels. Accordingly, the prior is computed as
\begin{equation}
\eprior{\stixel*{}} =
\sum_{\stixelidx=2}^{\numstixels}
\energyf[pair]{\stixeld*,\stixel*{\stixelidx-1}} +
\energyf[first]{\stixel*{1}} \punctspace .
\label{eq:prior}
\end{equation}
In this paper, we propose a new additional prior term that uses the specific properties of the three geometric classes. We expect the two random variables $\planea , \planeb$ representing the plane parameters of a Stixel to be Gaussian distributed, \ie
\begin{equation}
\edepthmodel{\stixeld*} = \left(\frac{ \planea* - \mu_{\classd*}^{\planea*}  }
                                     {\sigma_{\classd*}^{\planea*}}           \right)^2
                          +
                          \left(\frac{ \planeb* - \mu_{\classd*}^{\planeb*}  }
                                     {\sigma_{\classd*}^{\planeb*}}           \right)^2
                          - \func{\log}{\partitionfunction} \punctspace .
\end{equation}
This prior favors planes in accordance to the expected 3D layout corresponding to the geometric class.  \Eg \textit{object} Stixels are expected to have an approximately constant disparity, \ie $\mu_{\textit{object}}^{\planeb*} = 0$. The expected road slant $\mu_{\textit{ground}}^{\planea*}$ can be set using prior knowlede or a preceeding road surface detection. Note that the novel formulation is a strict generalization of the original method, since they are equivalent, if the slant is fixed, \ie $\sigma_{object}^{\planeb*} \rightarrow 0, \mu_{object}^{\planeb*} = 0$.

\subsection{Inference}
\label{sec:new_model}
The sophisticated energy function defined in \cref{sec:method} is optimized via Dynamic Programming as in \cite{Pfeiffer2011}. However, we also have to optimize jointly for the novel depth model. When optimizing for the plane parameters $\planea*_i,\planeb*_i$ of a certain Stixel $\stixeld*$, it becomes apparent that all other optimization parameters are independent of the actual choice of the plane parameters. We can thus simplify
\begin{align}
\argminx*{\planea*_i,\planeb*_i} \energyf{\stixel*{},\meas*{}} =
         \argminx*{\planea*_i,\planeb*_i} \energyf[stixel]{\stixeld*,\meas*{}} + \edepthmodel{\stixeld*}
        \punctspace .
\end{align}
Thus, we minimize the global energy function with respect to the plane parameters of all Stixels and all geometric classes independently. We can find an optimal solution of the
resulting weighted least squares problem in closed form, however, we still need to compare the Stixel disparities to our new plane depth model. Therefore, the complexity added to the original formulation is another quadratic term in the image height.

\begin{figure}[tb]
  \centering
      \begin{tabular}{@{}c@{}c@{}}
        \includegraphics[width=0.37\textwidth]{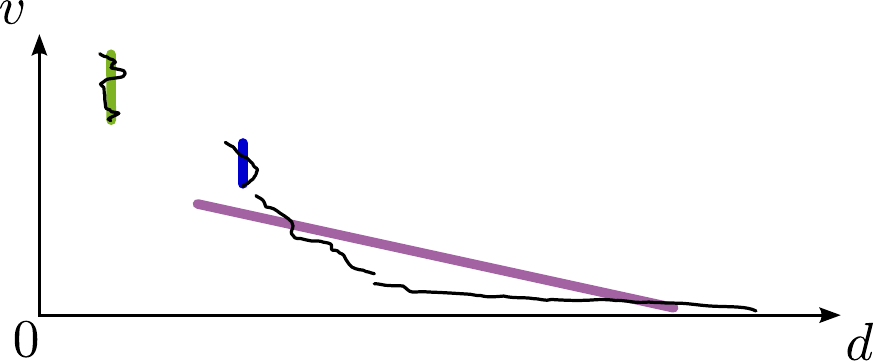} &
        \includegraphics[width=0.59\textwidth]{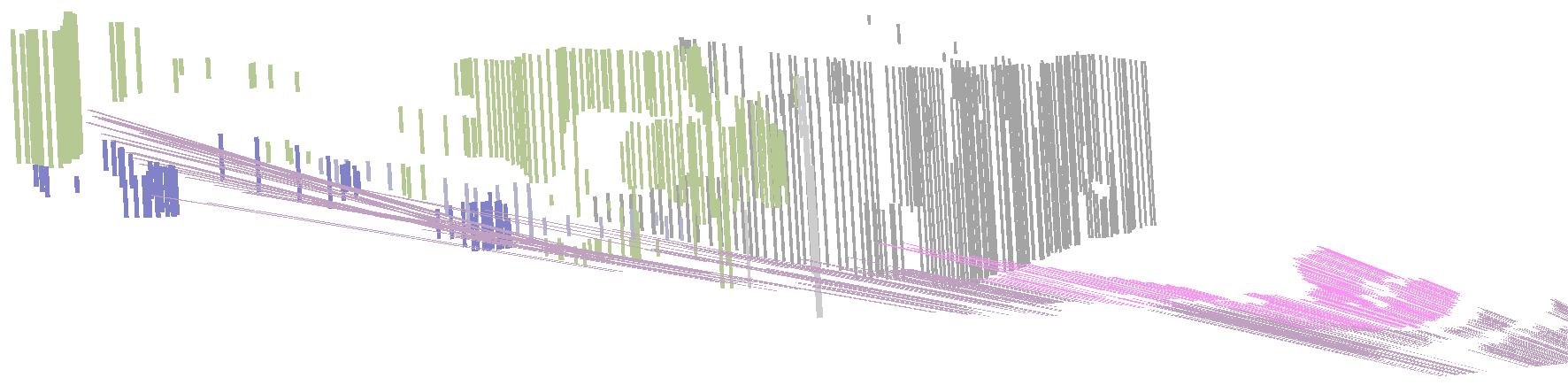} \\
        \includegraphics[width=0.37\textwidth]{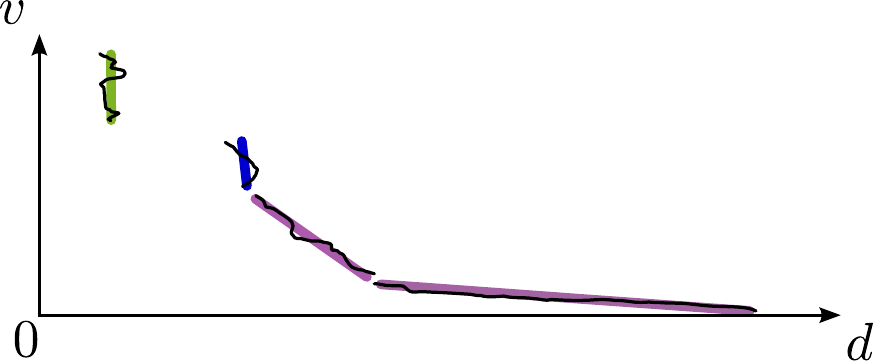} &
        \includegraphics[width=0.59\textwidth]{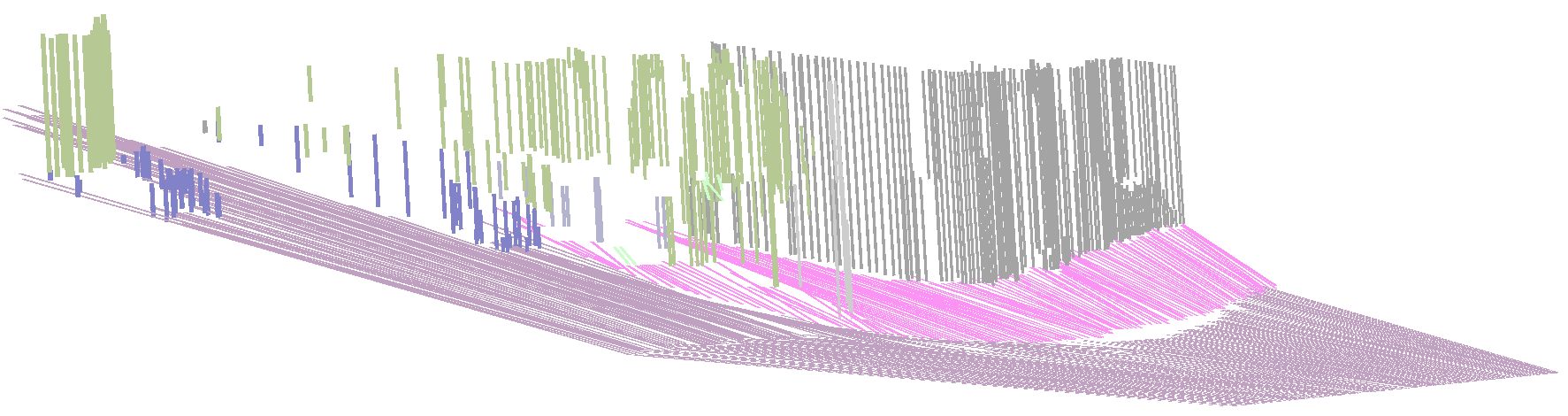}
      \end{tabular}
\captionsetup{skip=\dimexpr\abovecaptionskip+10pt}
\caption{Comparison of original \cite{Schneider2016} (top) and our slanted (bottom) Stixels: due to the
fixed slant in the original formulation the road
surface is not well represented as illustrated on the left.
The novel model is capable to reconstruct the whole scene accurately.
}

\end{figure}

\subsection{Stixel Cut Prior}
\label{subsec:cutprior}
The Stixel inference described so far requires to estimate the costs for each possible Stixel in an image, although many Stixels could be trivially discarded, \eg in image regions with homogeneous depth and semantic input. We propose a novel prior that can be easily used to significantly reduce the computational burden by exploiting hypothesis generation. To this end, we formulate a new prior similar to \cite{Cordts2014}, but instead of Stixel bottom and top probabilities we incorporate generic likelihoods for pixels being the cut between two Stixels.

We leverage this additional information adding a novel term for a Stixel $\stixeld*$
\begin{equation}
\energyf[cut]{\stixeld*} =
        \func{-\log}{\func{\cutpriorval*{\rowbotd*}}{cut}} \punctspace ,
\end{equation}
where $\func{\cutpriorval*{\rowbotd*}}{cut}$ is the confidence for a cut at $\rowbotd*$, thus $\func{\cutpriorval*{\rowbotd*}}{cut} = 0$ implies that there is no cut between two Stixels at row $\row$.

As described in \cite{PfeifferDiss}, we can use a recursive definition of the optimization problem to design the Dynamic Programming solution scheme. In order to simplify our description we use a special notation to refer to Stixels: $ob_{b}^{t} = \{v^{b}, v^{t}, object\}$. Similarly, $OB^k$ is defined as the minimum aggregated cost of the best segmentation from position $0$ to $k$. The Stixel at the end of the segmentation associated with each minimum cost is denoted as $ob^k$. We next show a recursive definition of the problem:
\begin{equation}
\begin{split}
OB^k = min \begin{cases} E_{data}(ob_0^k)+E_{prior}(ob_0^k)\\
E_{data}(ob_x^k)+E_{prior}(ob_x^k,ob^{x-1})+ OB^{x-1}  \forall x \in cuts, x \leq k \\
E_{data}(ob_x^k)+E_{prior}(ob_x^k,gr^{x-1})+ GR^{x-1}  \forall x \in cuts, x \leq k \\
E_{data}(ob_x^k)+E_{prior}(ob_x^k,sk^{x-1})+ SK^{x-1}  \forall x \in cuts, x \leq k
\end{cases}
\end{split}
\label{eq:dyn_programming}
\end{equation}
We only show the case for object Stixels, but the other cases are solved similarly. Also, $GR^k$ and $SK^k$ stand for ground and sky respectively. The base case problem, \ie segmenting a column of the single pixel at the bottom, is defined: $OB^0 = E_{data}(ob_0^0)+E_{prior}(ob_0^0)$. Our method trusts that all the optimal cuts will be included in our over-segmentation ($cuts$ in \cref{eq:dyn_programming}), therefore, only those positions are checked as Stixel start and end. This reduces the complexity of the Stixel estimation problem for a single column to $\mathcal{O}(h' \times h')$, where $h'$ is the number of over-segmentation cuts computed for this column, $h$ is image height and $h' \ll h$.

The computational complexity reduction becomes apparent in \cref{fig:stixel_graph}.
As stated in \cite{Cordts2017}, the inference problem can be interpreted as finding the shortest path in a directed acyclic graph. Our approach prunes all the vertices associated with the Stixel's top row not included according to the Stixel cut prior, \cf \cref{fig:pruned_stixel_graph}.


\newlength{\spNodeDist} 

\setlength{\spNodeDist}{2.5mm}

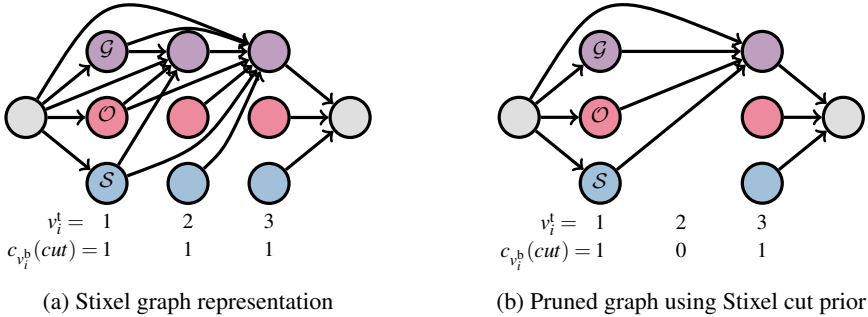
\begin{figure}
\footnotesize
\begin{subfigure}[t]{0.50\textwidth}
\centering
\begin{tikzpicture}


\tikzset{math mode/.style = { %
        execute at begin node=$, %
        execute at end node=$ %
    }}
\tikzset{math mode small/.style = { %
        execute at begin node=$\scriptstyle, %
        execute at end node=$ %
        }}

\tikzstyle{state}  = [ circle,
                       draw=black,
                       inner sep=0.2pt,
                       minimum size=15pt,
                       math mode,
                       node distance=1.2\spNodeDist and 2\spNodeDist,
                       text height=1.ex,
                       text depth=.25ex,
                       fill=gray!25,
                       pattern color=gray!25,
                       line width=1.2pt
                     ]

\definecolor{supportC}{RGB}{128, 64,128}
\definecolor{verticalC}{RGB}{220, 20, 60}
\definecolor{skyC}{RGB}{ 70,130,180}

\tikzstyle{supportN}  = [ state ,
                          fill=supportC!50
                        ]
\tikzstyle{verticalN} = [ state ,
                          fill=verticalC!50
                        ]
\tikzstyle{skyN}      = [ state ,
                          fill=skyC!50
                        ]

\tikzstyle{dummy}     = [ state ,
                          fill=none,
                          draw=none
                        ]

\tikzstyle{info}      = [ dummy,
                          node distance=-0.1\spNodeDist,
                        ]

\tikzstyle{info2}      = [ dummy,
                          node distance=-0.8\spNodeDist,
                        ]

\tikzstyle{trans}     = [ ->,
                          line width=1.2pt
                        ]



\node[state                     ]  (source)  {};

\node[verticalN, right=of source]  (v0)      {\object};
\node[supportN , above=of v0    ]  (s0)      {\ground};
\node[skyN     , below=of v0    ]  (k0)      {\sky};

\node[verticalN, right=of v0    ]  (v1)      {};
\node[supportN , above=of v1    ]  (s1)      {};
\node[skyN     , below=of v1    ]  (k1)      {};

\node[verticalN, right=of v1    ]  (vh)      {};
\node[supportN , above=of vh    ]  (sh)      {};
\node[skyN     , below=of vh    ]  (kh)      {};

\node[state    , right=of vh    ]  (sink)    {};

\node[info     , below=of k0    ]  (row1)    {1};
\node[info     , below=of k1    ]  (row2)    {2};
\node[info     , below=of kh    ]  (rowh)    {3};

\node[info2    , below=of row1  ]  (cut1)    {1};
\node[info2    , below=of rowh  ]  (cuth)    {1};
\node[info2    , below=of row2  ]  (cut2)    {1};

\node[info     , left =of row1  ]  (row)     {\rowtopd*=};
\node[info2    , left =of cut1  ]  (cut)     {\func{\cutpriorval*{\rowbotd*}}{cut}=};


\draw[trans] (source) -- (v0);
\draw[trans] (source) -- (s0);
\draw[trans] (source) -- (k0);
\draw[trans] (source) -- (s1);
\draw[trans] (source) .. controls ($(s0) + (-0,  3.5\spNodeDist)$) .. (sh);

\draw[trans] (v0)     -- (s1);
\draw[trans] (s0)     -- (s1);
\draw[trans] (k0)     -- (s1);

\draw[trans] (v0)     -- (sh);
\draw[trans] (s0)     .. controls ($(s1) + (0, 1.5\spNodeDist)$) .. (sh);
\draw[trans] (k0)     .. controls ($(v1) + (0.2, -1.6\spNodeDist)$) .. (sh);

\draw[trans] (v1)     -- (sh);
\draw[trans] (s1)     -- (sh);
\draw[trans] (k1)     .. controls ($(v1) + (0.5, -1.6\spNodeDist)$) .. (sh);

\draw[trans] (vh)     -- (sink);
\draw[trans] (sh)     -- (sink);
\draw[trans] (kh)     -- (sink);

\end{tikzpicture}
\captionsetup{skip=\dimexpr\abovecaptionskip-10pt}
\caption{Stixel graph representation}
\label{fig:stixel_graph}
\end{subfigure}
\begin{subfigure}[t]{0.50\textwidth}
\centering
\begin{tikzpicture}


\tikzset{math mode/.style = { %
        execute at begin node=$, %
        execute at end node=$ %
    }}
\tikzset{math mode small/.style = { %
        execute at begin node=$\scriptstyle, %
        execute at end node=$ %
        }}

\tikzstyle{state}  = [ circle,
                       draw=black,
                       inner sep=0.2pt,
                       minimum size=15pt,
                       math mode,
                       node distance=1.2\spNodeDist and 2\spNodeDist,
                       text height=1.25ex,
                       text depth=.25ex,
                       fill=gray!25,
                       pattern color=gray!25,
                       line width=1.2pt
                     ]

\definecolor{supportC}{RGB}{128, 64,128}
\definecolor{verticalC}{RGB}{220, 20, 60}
\definecolor{skyC}{RGB}{ 70,130,180}

\tikzstyle{supportN}  = [ state ,
                          fill=supportC!50
                        ]
\tikzstyle{verticalN} = [ state ,
                          fill=verticalC!50
                        ]
\tikzstyle{skyN}      = [ state ,
                          fill=skyC!50
                        ]

\tikzstyle{dummy}     = [ state ,
                          fill=none,
                          draw=none
                        ]

\tikzstyle{info}      = [ dummy,
                          node distance=-0.1\spNodeDist,
                        ]

\tikzstyle{info2}      = [ dummy,
                          node distance=-0.8\spNodeDist,
                        ]

\tikzstyle{trans}     = [ ->,
                          line width=1.2pt
                        ]



\node[state                     ]  (source)  {};

\node[verticalN, right=of source]  (v0)      {\object};
\node[supportN , above=of v0    ]  (s0)      {\ground};
\node[skyN     , below=of v0    ]  (k0)      {\sky};

\node[dummy    , right=of v0    ]  (dummyV)  {};
\node[dummy    , right=of s0    ]  (dummyS)  {};
\node[dummy    , right=of k0    ]  (dummyK)  {};

\node[verticalN, right=of dummyV]  (vh)      {};
\node[supportN , above=of vh    ]  (sh)      {};
\node[skyN     , below=of vh    ]  (kh)      {};

\node[state    , right=of vh    ]  (sink)    {};

\node[info     , below=of k0     ]  (row1)    {1};
\node[info     , below=of kh     ]  (rowh)    {3};
\node[info     , below=of dummyK ]  (dot2)    {2};

\node[info2     , below=of row1  ]  (cut1)    {1};
\node[info2     , below=of rowh  ]  (cuth)    {1};
\node[info2     , below=of dot2  ]  (cut2)    {0};

\node[info     , left =of row1  ]  (row)     {\rowtopd*=};
\node[info2    , left =of cut1  ]  (cut)     {\func{\cutpriorval*{\rowbotd*}}{cut}=};

\draw[trans] (source) -- (v0);
\draw[trans] (source) -- (s0);
\draw[trans] (source) -- (k0);
\draw[trans] (source) .. controls ($(s0) + (-0,  3.5\spNodeDist)$) .. (sh);
\draw[trans] (v0)     -- (sh);
\draw[trans] (s0)     -- (sh);
\draw[trans] (k0)     -- (sh);

\draw[trans] (vh)     -- (sink);
\draw[trans] (sh)     -- (sink);
\draw[trans] (kh)     -- (sink);

\end{tikzpicture}
\captionsetup{skip=\dimexpr\abovecaptionskip-10pt}
\caption{Pruned graph using Stixel cut prior}
\label{fig:pruned_stixel_graph}
\end{subfigure}
\captionsetup{skip=\dimexpr\abovecaptionskip+10pt}
\caption{Stixel inference illustrated as shortest path problem on a directed acyclic graph: the Stixel segmentation is computed by finding the shortest path from the source (left gray node) to the sink (right gray node).
The vertices represent Stixels with colors encoding their geometric class, \ie \textbf{g}round, \textbf{o}bject and \textbf{s}ky.
Only the incoming edges of ground nodes are shown for simplicity. Adapted from~\cite{Cordts2017}.
}
\end{figure}


\section{Experiments}
\label{sec:experiments}
This section assesses the accuracy and run-time of our proposal. A previous concern is to verify that our method maintains the accuracy for scenes containing only flat roads and represents non-flat roads better. For that purpose, we evaluate on both synthetic and real data, \cf \cref{subsec:datasets}. We introduce metrics and baselines as well as a Stixel over-segmentation method and other details in \cref{subsec:experiment_details}. Finally, quantitative and qualitative results are reported, \cf \cref{subsec:results}.

\subsection{Datasets}
\label{subsec:datasets}
As our Stixel model represents geometric and semantic information, we measure the accuracy of our method for both. Therefore, we evaluate on the -to the best of our knowledge- only such dataset: an annotated subset of KITTI \cite{Ladicky2014} called $Ladicky$. It consists of 60 images with a resolution of 0.5 MP that we all use for evaluation. We follow the suggestion of the original author to ignore the three rarest object classes, leaving a set of 8 classes. Following \cite{Schneider2016}, we use additional publicly available semantic annotations on other parts of KITTI for training. All in all, we have a training set of 676 images, where we harmonized the object classes used by the different authors.

We also evaluate disparity accuracy on the training data of the well-known stereo challenge KITTI 2015 \cite{Geiger2012CVPR}. This dataset comprises a set of 200 images with sparse disparity ground truth obtained from a laser scanner. However, there is no suitable semantic ground truth available for this dataset.
Furthermore, we evaluate the semantic accuracy on Cityscapes \cite{Cordts2016}, a highly complex dataset with dense annotations of 19 classes on ca. 3000 images for training and 500 images for validation that we used for testing.

\begin{figure}[bt]
  \centering
  \begin{subfigure}[t]{0.32\textwidth}
    \centering
    \includegraphics[trim=100 200 250 100,clip,width=1\linewidth]{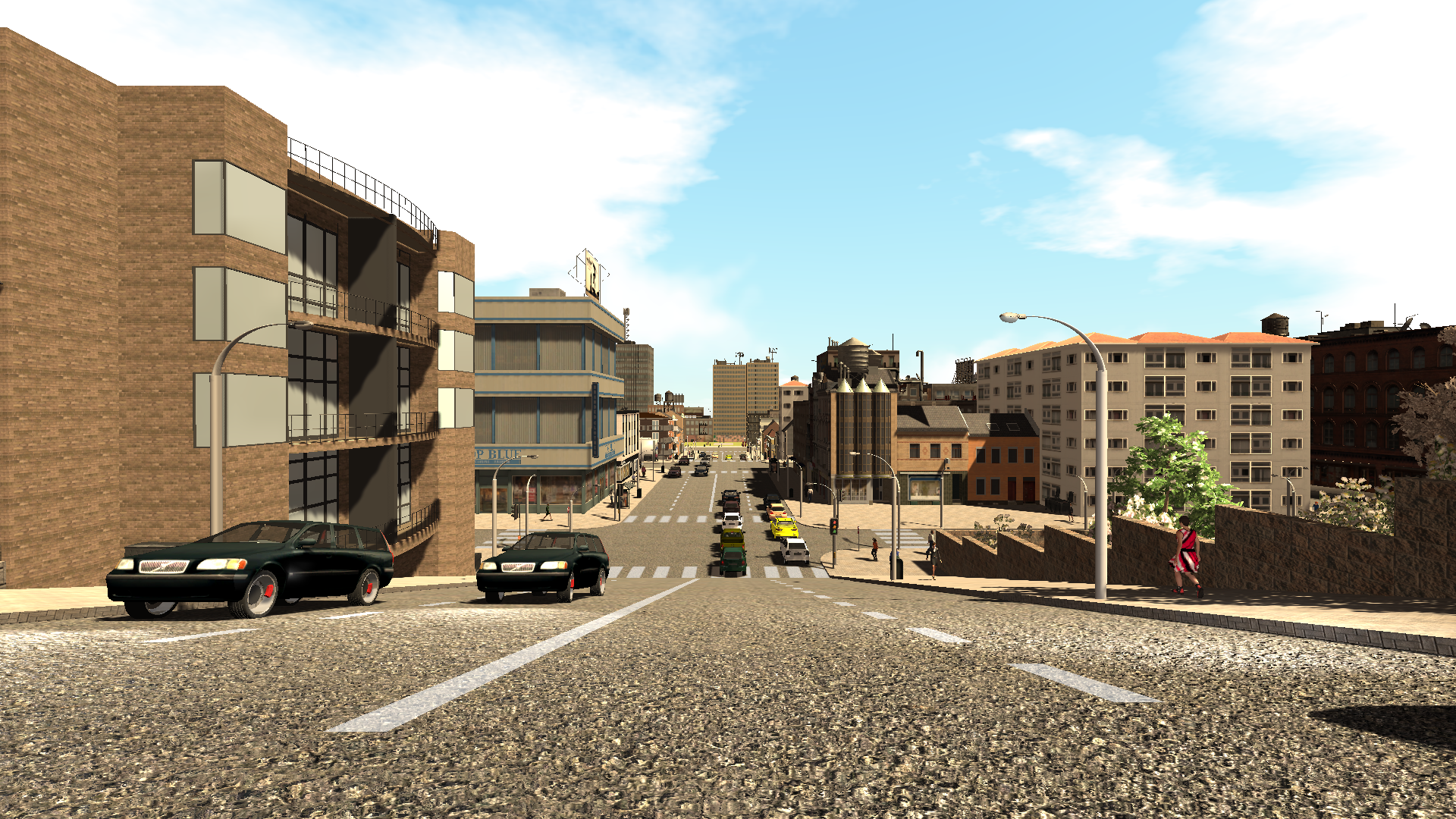}
  \end{subfigure}
  \begin{subfigure}[t]{0.33\textwidth}
    \centering
    \includegraphics[trim=100 200 250 100,clip,width=1\linewidth]{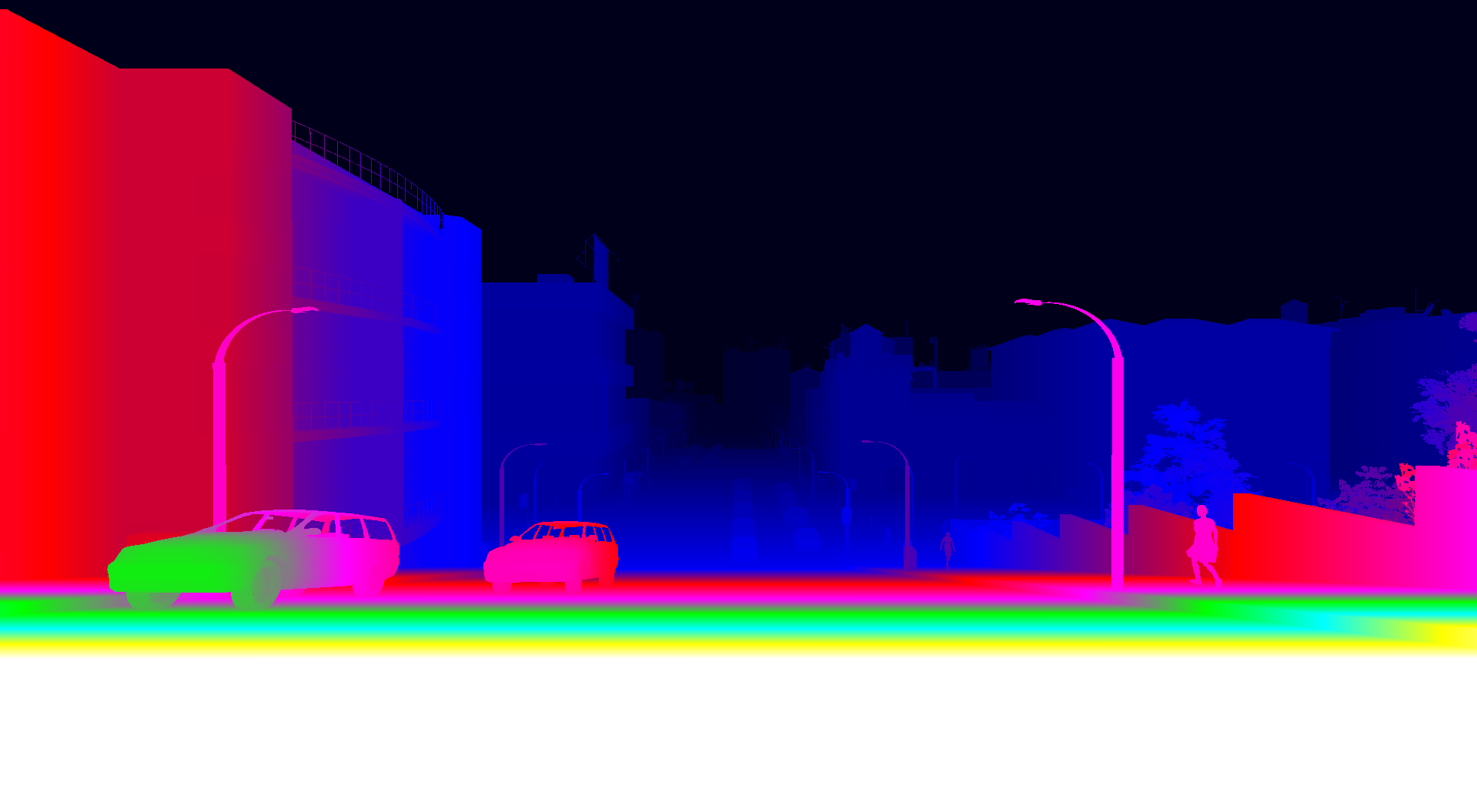}
  \end{subfigure}
  \begin{subfigure}[t]{0.32\textwidth}
    \centering
    \includegraphics[trim=100 200 250 100,clip,width=1\linewidth]{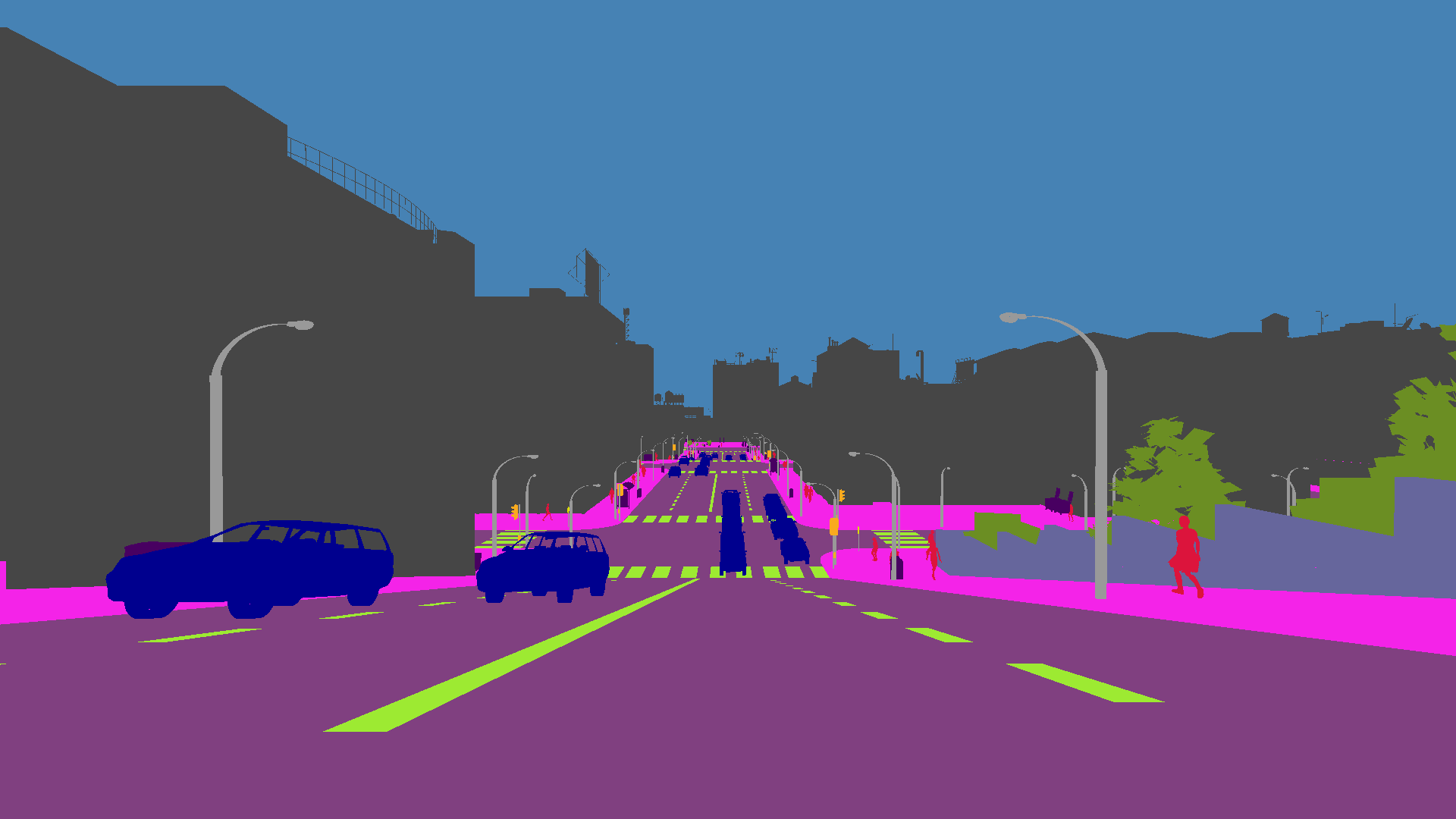}
  \end{subfigure}
  \begin{subfigure}[t]{\textwidth}
  \begin{tabularx}{0.99\textwidth}{XXXXXXXX}
    \labelcolorb{sidewalk}      &
    \labelcolor{building} &
    \labelcolor{vegetation}    &
    \labelcolorb{traffic light} &
    \labelcolorb{traffic sign}  &
    \labelcolor{bicycle} &
    \labelcolor{motorcycle} &
    \labelcolorb{road lines}
  \end{tabularx}
  \begin{tabularx}{0.99\textwidth}{XXXXXXXXXXXX}
    \labelcolorb{terrain} &
    \labelcolor{road}          &
    \labelcolor{wall} &
    \labelcolor{pole} &
    \labelcolorb{rider} &
    \labelcolor{truck} &
    \labelcolor{bus} &
    \labelcolor{train} &
    \labelcolorb{fence} &
    \labelcolor{person}        &
    \labelcolor{sky} &
    \labelcolor{car}
  \end{tabularx}%
  \end{subfigure}
\captionsetup{skip=\dimexpr\abovecaptionskip+10pt}
\caption{The \datasetname{} Dataset. A sample frame (left) with its depth (center) and semantic labels (right).}
\label{fig:synthia-sf}
\end{figure}

Unfortunately, all the above datasets were generated in flat road environments, and they only help us validate that we are not decreasing our accuracy for this case. In order to compare the accuracy of competing algorithms on non-flat road scenarios, a new dataset is required.
Therefore, we introduce a new synthetic dataset inspired by \cite{RosCVPR16}. This dataset has been generated with the purpose of evaluating our model, but it contains enough information to be useful in additional related tasks, such as object recognition, semantic and instance segmentation, among others.
\datasetnamefull{} (\datasetname) consists of photo-realistic frames rendered from a virtual city and comes with precise pixel-level depth and semantic annotations for 19 classes (see \cref{fig:synthia-sf}). This new dataset contains 2224 images that we use to evaluate both depth and semantic accuracy.

\subsection{Experiment Details}
\label{subsec:experiment_details}
We evaluate our proposed method in terms of semantic and depth accuracy using two \textbf{metrics}. The depth accuracy is obtained as the outlier rate of the disparity estimates, the standard metric used to evaluate on KITTI benchmark \cite{Geiger2012CVPR}. An outlier is a disparity estimation with an absolute error larger than 3 px or a relative deviation larger than 5\% compared to the ground truth. The semantic accuracy is evaluated with average Intersection-over-Union (IoU) over all classes, also a standard measure for semantics \cite{Everingham2015}. We also provide Frame-rate (Hz) to ensure our system is capable of real-time performance and number of Stixels per image to quantify the complexity of the obtained representation.
All Stixel run-times are obtained using a multi-threaded implementation on standard consumer hardware: Intel i7-6800K.
For the FCN output, a Maxwell NVidia Titan X is used.

Semantic Stixels \cite{Schneider2016} serves as \textbf{baseline}, because they achieve state-of-the-art results in terms of Stixel accuracy. We provide the accuracy of our new disparity model, \cf \cref{sec:method}. Finally, we also evaluate our reduced complexity approach, \cf \cref{subsec:cutprior}, both for our model and Semantic Stixels, \textit{Ours (fast)} and \textit{\cite{Schneider2016} (fast)}, respectively.

As \textbf{input}, we use disparity images $\disp{}$ obtained via semi-global-matching (SGM) \cite{Hirschmuller2008} and pixel-level semantic labels $\pixclassval{}$ computed by a fully convolutional network (FCN) \cite{Long2015}. We use the same FCN model used in \cite{Schneider2016} without retraining to allow for comparison.
For the same reason, we set Stixel width to 8 px.
The same downscaling is applied in the v-direction. The rest of the parameters are taken from \cite{Schneider2016}.
We use the known camera calibration to obtain expected $\mu_{ground}^{\planea*}$ and $\mu_{ground}^{\planeb*}$. For objects, we set $\sigma_{object}^{\planeb*} \rightarrow 0, \mu_{object}^{\planeb*} = 0$ because the disparity is too noisy for the slanted object model. Finally, for Sky Stixels it does not make sense to have slanted surfaces, therefore, we set: $\mu_{sky}^{\planea*}=0, \mu_{sky}^{\planea*}=0, \sigma_{sky}^{\planea*} \rightarrow 0, \sigma_{sky}^{\planeb*} \rightarrow 0$.

In order to improve the efficiency of our approach, we use a preceding \textbf{Stixel cut prior generation}. Our goal is to show that we can speed up the Stixel computation with a very simple and fast approach. Accordingly, we opt for \cite{Oana2016} to generate our over-segmentation: a very low-level method based on simple mathematical concepts like the use of local and strict extrema of the disparity map to find points of interest. This method first performs an \textit{Extreme points detection} step that generates possible Stixel cuts and subsequently filters them. As we are only interested in cut hypothesis we only use the first step. We also use semantic segmentation to generate cuts, when there is a change in semantic class we assume there is a cut.

\begin{figure}[bt]
\centering
\begin{subfigure}[t]{0.30\textwidth}
  \caption*{Left Image}
  \centering
  \includegraphics[trim= 0 0 0 0.2cm, width=1\linewidth,height=1.7cm,keepaspectratio]{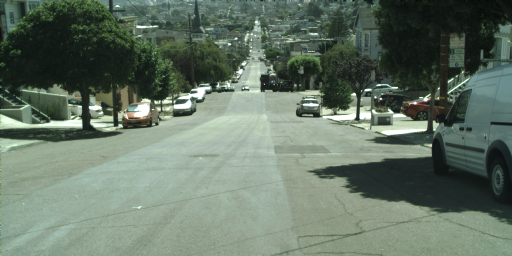}
\end{subfigure}
\begin{subfigure}[t]{0.30\textwidth}
  \caption*{Original Stixels \cite{Schneider2016}}
  \centering
  \includegraphics[width=1\linewidth,height=1.7cm,keepaspectratio]{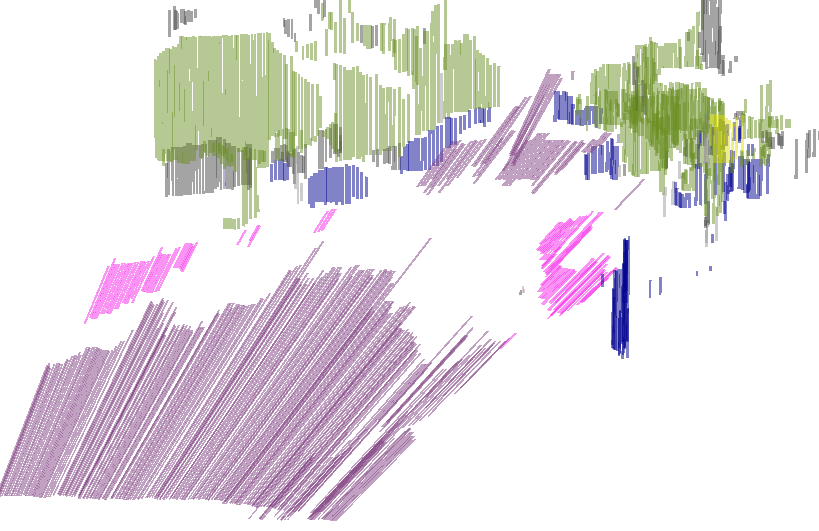}
\end{subfigure}
\begin{subfigure}[t]{0.30\textwidth}
  \caption*{Our Stixels}
  \centering
  \includegraphics[width=1\linewidth,height=1.7cm,keepaspectratio]{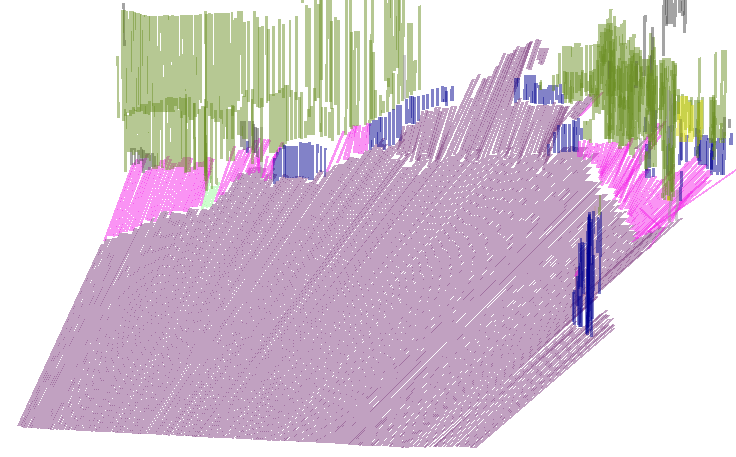}
\end{subfigure}

\begin{subfigure}[t]{0.30\textwidth}
  \centering
  \includegraphics[trim= 0 0 0 0.2cm, width=1\linewidth,height=1.7cm,keepaspectratio]{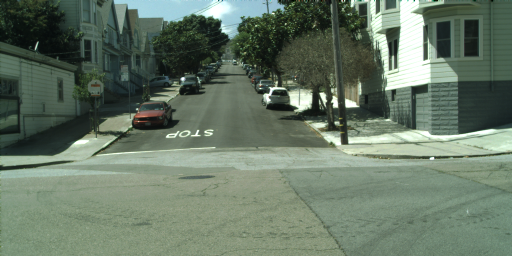}
\end{subfigure}
\begin{subfigure}[t]{0.30\textwidth}
  \centering
  \includegraphics[width=1\linewidth,height=1.7cm,keepaspectratio]{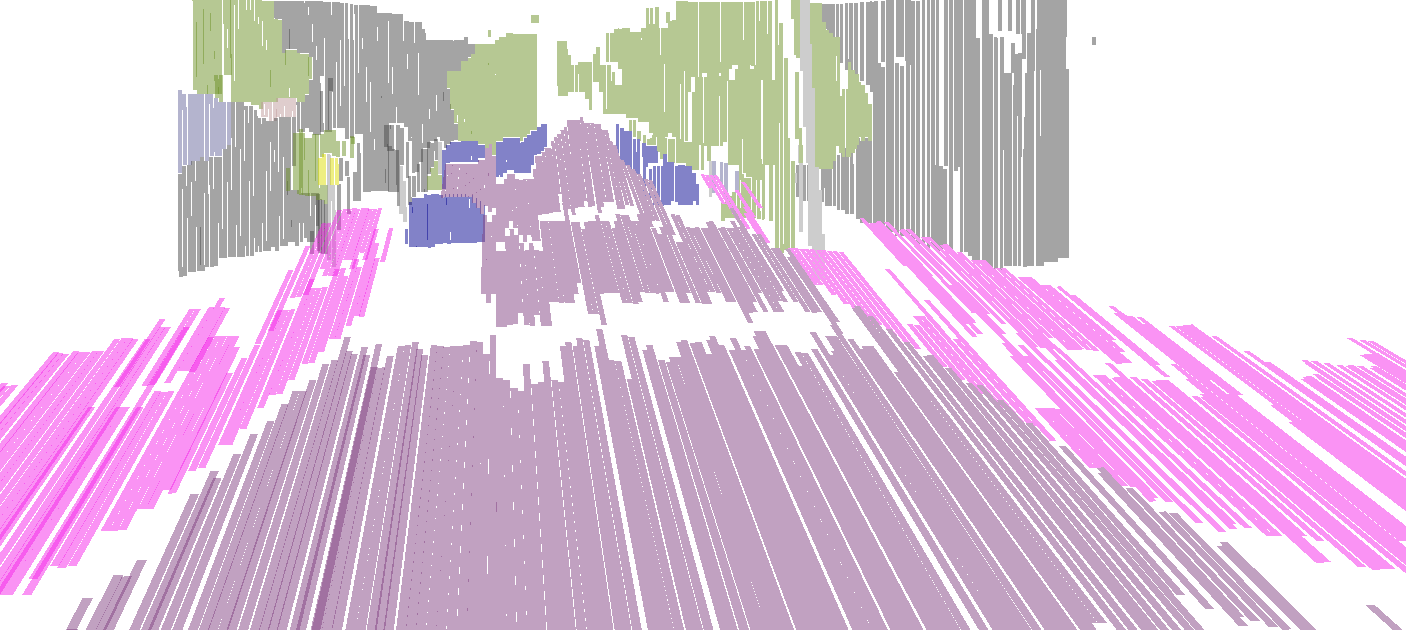}
\end{subfigure}
\begin{subfigure}[t]{0.30\textwidth}
  \centering
  \includegraphics[width=1\linewidth,height=1.7cm,keepaspectratio]{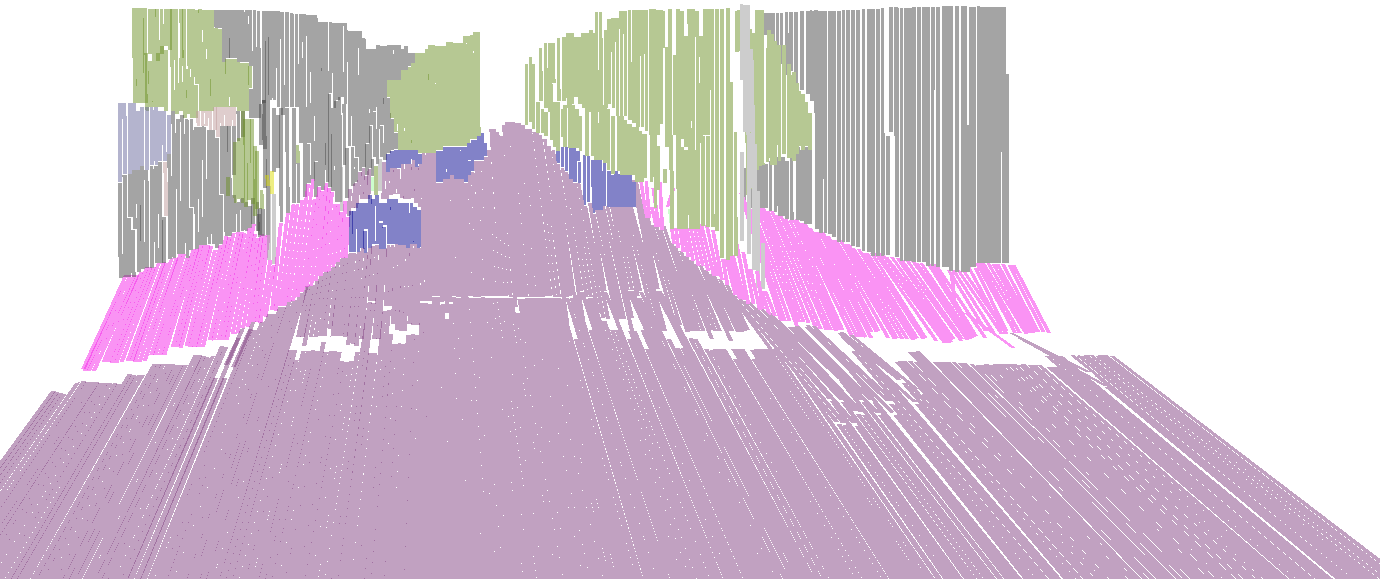}
\end{subfigure}

\begin{subfigure}[t]{0.30\textwidth}
  \centering
  \includegraphics[trim= 0 0 0 0.2cm, width=1\linewidth,height=1.7cm,keepaspectratio]{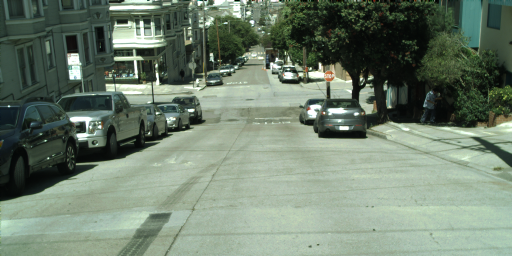}
\end{subfigure}
\begin{subfigure}[t]{0.30\textwidth}
  \centering
  \includegraphics[width=1\linewidth,height=1.7cm,keepaspectratio]{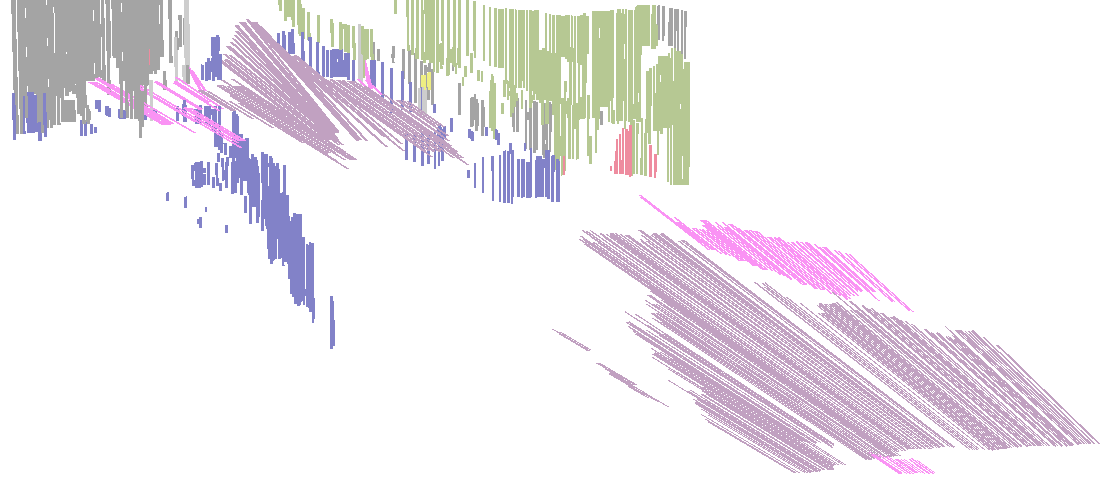}
\end{subfigure}
\begin{subfigure}[t]{0.30\textwidth}
  \centering
  \includegraphics[width=1\linewidth,height=1.7cm,keepaspectratio]{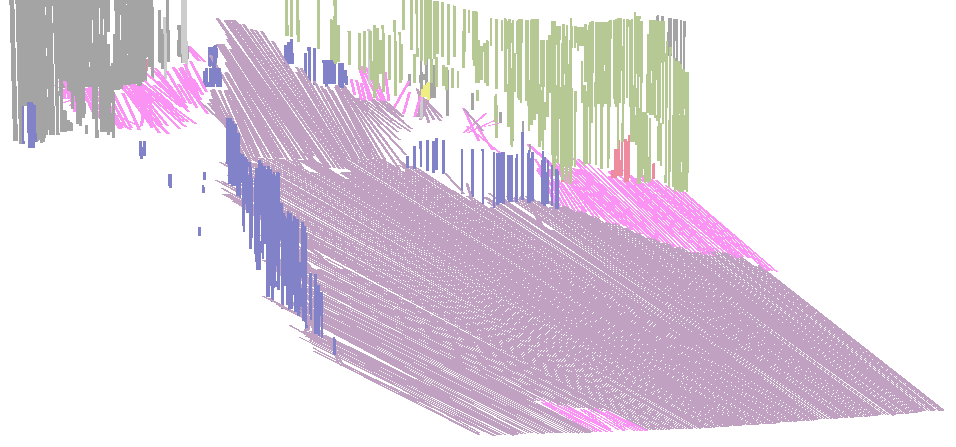}
\end{subfigure}

\begin{subfigure}[t]{0.30\textwidth}
  \centering
  \includegraphics[trim= 0 0 0 0.2cm, width=1\linewidth,height=1.7cm,keepaspectratio]{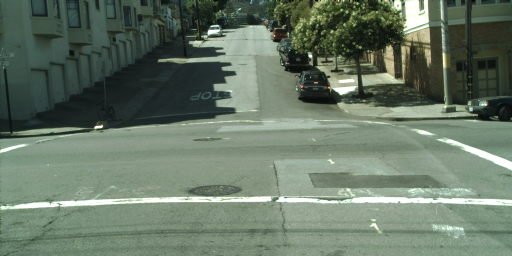}
\end{subfigure}
\begin{subfigure}[t]{0.30\textwidth}
  \centering
  \includegraphics[width=1\linewidth,height=1.7cm,keepaspectratio]{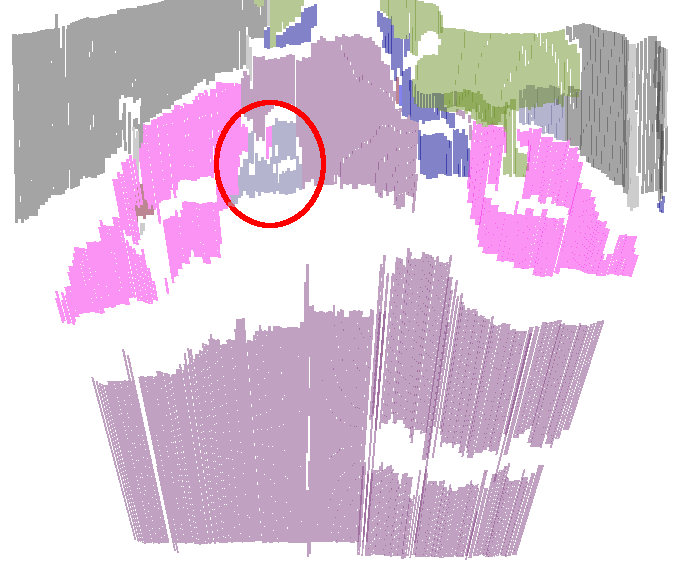}
\end{subfigure}
\begin{subfigure}[t]{0.30\textwidth}
  \centering
  \includegraphics[width=1\linewidth,height=1.7cm,keepaspectratio]{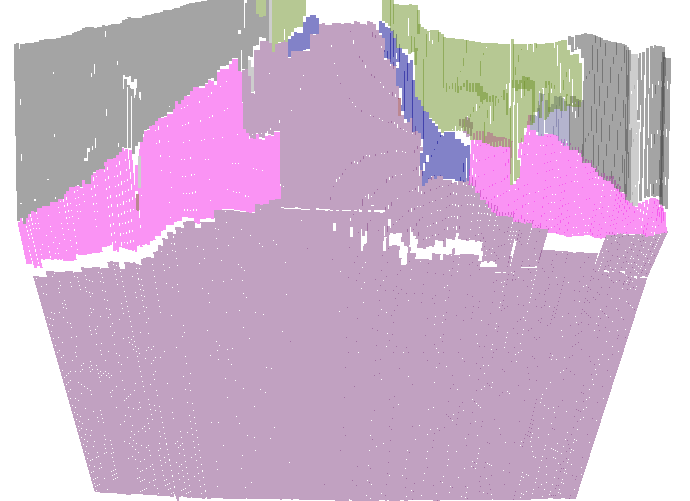}
\end{subfigure}

\captionsetup{skip=\dimexpr\abovecaptionskip+10pt}
\caption{Exemplary outputs on real data: in all scenes with non-flat roads our model correctly represents the scene, while retaining accuracy on objects. The last line shows a failure case, where our approach classifies the road as sidewalk due to erroneous semantic input. However, the original approach reconstructs a wall in this case, highlighted by a red circle. This could lead \eg to an emergency break.}
\label{fig:comparison}
\end{figure}

\subsection{Results}
\label{subsec:results}

The quantitative results of our proposals and baselines as described in \cref{sec:method} are shown in \cref{table:results}.
The first observation is that all variants are compact representations of the surrounding, since the complexity of the Stixel representation is small compared to the high resolution input images, \cf the last row in the \cref{table:results}.

Second, our method achieves comparable or slightly better results on all datasets with flat roads. This indicates that the novel and more flexible model does not harm the accuracy in such scenarios.

The third observation is that the proposed \textit{fast} variant improves the run-time of both the original Stixel approach by up to 2x and the novel Stixel approach by up to 7x with only a slight drop in depth accuracy. The benefit increases with higher resolution input images.
This is due to the mean density of Stixel cuts in our over-segmentation for SYNTHETIC-SF of 13\% with standard deviation of 2, which is equivalent to a 8x reduced vertical resolution.

Finally, we observe that our novel model is able to accurately represent non-flat scenarios in contrast to the original Stixel approach yielding a substantially increased depth accuracy by more than $17\%$. We also improve in terms of semantic accuracy, which we address to the joint semantic and depth inference that benefits of a better depth model.

The observations from the quantitative evaluation are confirmed also in the qualitative results, \cf \cref{fig:comparison}.

\begin{table}[bt]
\footnotesize
\centering
\captionsetup{skip=\dimexpr\abovecaptionskip+10pt}
\caption{Quantitative results of our methods compared to \cite{Schneider2016}, raw SGM and FCN. Significantly best results highlighted in bold.}
\begin{tabular}{ll|ll|llll}
\toprule
Metric                            & Dataset         & SGM   & FCN  & \cite{Schneider2016} & Ours & \cite{Schneider2016} (fast) & Ours (fast)  \\
\midrule
\multirow{3}{*}{Disp Error (\%)}  & Ladicky         & 16.6  & -    & 17.3      & 16.9             & 18.5 & 17.8            \\
                                  & KITTI 15        & 11.5  & -    & 10.9      & 11.0             & 11.8 & 11.7            \\
                                  & \datasetname    & 11.0  & -    & 30.9      & \textbf{12.9}    & 33.9 & 15.4            \\

\hline

\multirow{3}{*}{IoU (\%)}         & Ladicky         & -     & 69.8 & 63.5      & 63.4             & 63.9  & 63.7           \\
                                  & Cityscapes      & -     & 60.8 & 65.7      & 65.8             & 65.7  & 65.8           \\
                                  & \datasetname    & -     & 45.9 & 46.0      & \textbf{48.5}    & 46.9  & \textbf{48.5}  \\

\hline

\multirow{3}{*}{Frame-rate (Hz)}  & KITTI           & 55    & 47.6 & 113       & 61               & 120   & 116            \\
                                  & Cityscapes      & 22    & 15.4 & 20.9      & 6.6              & 36.6  & 27.5           \\
                                  & \datasetname    & 21    & 13.9 & 19.4      & 4.7              & 38.9  & 33.1           \\

\hline

\multirow{3}{*}{\# Stixels ($10^3$)} & KITTI        & 226   & 226  & 0.6       & 0.6              & 0.6   & 0.6            \\
                                     & Cityscapes   & 1 k   & 1 k  & 1.4       & 1.5              & 1.3   & 1.4            \\
                                     & \datasetname & 1 k   & 1 k  & 1.5       & 1.7              & 1.2   & 1.3            \\
\bottomrule
\end{tabular}
\label{table:results}
\end{table}

\section{Conclusions}
\label{sec:conclusions}

This paper presented a novel depth model for the Stixel World that is able to represent non-flat roads and slanted objects in a compact representation that overcomes the previous restrictive constant height and depth assumptions respectively. Moreover, a novel approximation is introduced in order to reduce the computational complexity significantly by only checking reasonable Stixel cuts. This representation change is required for difficult environments that are found in the real world. We showed in extensive experiments on several related datasets that our depth model is able to better represent those scenes and our approximation is able to reduce the run-time drastically with only a slight drop in accuracy.

\section{Acknowledgements}
\label{sec:acks}

This research has been supported by the MICINN under contract number TIN2014-53234-C2-1-R. By the MEC under contract number TRA2014-57088-C2-1-R and the Generalitat de Catalunya projects 2014-SGR-1506 and 2014-SGR-1562, we also thank CERCA Programme / Generalitat de Catalunya, NVIDIA for the donation of the systems used in this work and SEBAP for the internship funding program. Finally, we thank Francisco Molero, Marc Garc{\'i}a, and the SYNTHIA team for the dataset generation.

\bibliography{egbib}
\end{document}